\documentclass[runningheads]{llncs}
\usepackage[T1]{fontenc}
\usepackage{graphicx}
\usepackage{multirow}
\usepackage{amsmath}
\usepackage[most]{tcolorbox}

\usepackage{xcolor}
\definecolor{purpleColor}{HTML}{A349A4}  
\definecolor{greenColor}{HTML}{3CA043}
\definecolor{orangeColor}{HTML}{FF8C42}
\begin{document}
\title{RALLM-POI: Retrieval-Augmented LLM for Zero-shot Next POI Recommendation with Geographical Reranking}
\titlerunning{Retrieval Augmented LLM for Zero-shot Next POI Recommendation}
%
\author{Kunrong Li \and
Kwan Hui Lim}
%
\authorrunning{K. Li and K. H. Lim}
%
\institute{Singapore University of Technology and Design, Singapore \\\email{kunrong\_li@mymail.sutd.edu.sg, kwanhui\_lim@sutd.edu.sg}}
%
\maketitle              
\begin{abstract}

Next point-of-interest (POI) recommendation predicts a user’s next destination from historical movements. Traditional models require intensive training, while LLMs offer flexible and generalizable zero-shot solutions but often generate generic or geographically irrelevant results due to missing trajectory and spatial context. To address these issues, we propose RALLM-POI, a framework that couples LLMs with retrieval-augmented generation and self-rectification. We first propose a Historical Trajectory Retriever (HTR) that retrieves relevant past trajectories to serve as contextual references, which are then reranked by a Geographical Distance Reranker (GDR) for prioritizing spatially relevant trajectories. Lastly, an Agentic LLM Rectifier (ALR) is designed to refine outputs through self-reflection. Without additional training, RALLM-POI achieves substantial accuracy gains across three real-world Foursquare datasets, outperforming both conventional and LLM-based baselines. Code is released at \url{https://github.com/LKRcrocodile/RALLM-POI}.

\end{abstract}
\section{Introduction}

Next Point-of-Interest (POI) recommendation aims to predict a user’s next destination from historical trajectories containing rich spatial-temporal patterns \cite{lim2019toursurvey}. This task benefits not only Location-Based Social Networks (LBSNs) but also broader domains such as traffic management, urban planning, and public health. Deep learning models including RNNs (LSTM, GRU), graph-based methods, and more recently transformers have achieved strong results by capturing sequential, structural, and contextual dependencies \cite{huang2019attention,liao2018mrnn,wang2024dyaglgraph2,vaswani2017attention,lin2021clte,zhang2022cfprec,duan2023clsprec}. However, these approaches suffer from cold-start and data sparsity problems, limiting their effectiveness in real-world applications.

Recently, LLMs have demonstrated significant advances \cite{li2024semantic,feng2024llmmove,llmmob,llmzs} in recommendation tasks. 
Some research has begun to explore the integration of LLMs into POI recommendation systems via a zero-shot manner. 
For example, LLMMove \cite{feng2024llmmove} and LLM-Mob \cite{llmmob} adopt delicate designed prompts to feed individual users’ check-in histories into pre-trained LLMs.
LMZS \cite{llmzs} assessed several pre-trained LLMs by providing LLMs with
examples of input and expected outputs to enhance predictive performance.
However, existing zero-shot LLM-based approaches also have notable limitations. 
Simply using prompt engineering with recent histories can be suboptimal, as the selected information may be less informative or only provide irrelevant check-ins, potentially distracting the LLM. 
This motivates us to design an approach that ensures that the most relevant trajectory information is provided to the LLM.

To address these issues, we propose a \textbf{R}etrieval-\textbf{A}ugmented \textbf{L}arge \textbf{L}anguage \textbf{M}odel for next \textbf{POI} recommendation, termed \textbf{RALLM-POI}. Specifically, it first applies a Historical Trajectory Retriever (HTR) that identifies highly similar historical user trajectories from a database, and prompt the LLM with contextually rich and personalized information. To enhance spatial plausibility, we propose the Geographical Distance Reranker (GDR), which prioritizes supporting trajectories based on dynamic spatial alignment using Decaying Weighted Dynamic Time Warping (DWDTW). By taking into account the user’s recent geographic movements and introducing recency-biased weights, GDR reranks the retrieved examples to better align with realistic spatial dynamics. Finally, after the LLM generates its recommendations, we introduce Agentic LLM Rectifier (ALR) that serves as a quality assurance layer to verify and rectify the response of the prior LLM response, which ensures that the final recommendations are robust and adhere strictly to task requirements. We extensively validate our RALLM-POI on three real-world Foursquare datasets, our method surpasses prior training-based and zero-shot LLM based methods remarkably.

\section{Methodology}

\begin{figure}[t]
  \centering
    \includegraphics[width=0.95\linewidth]{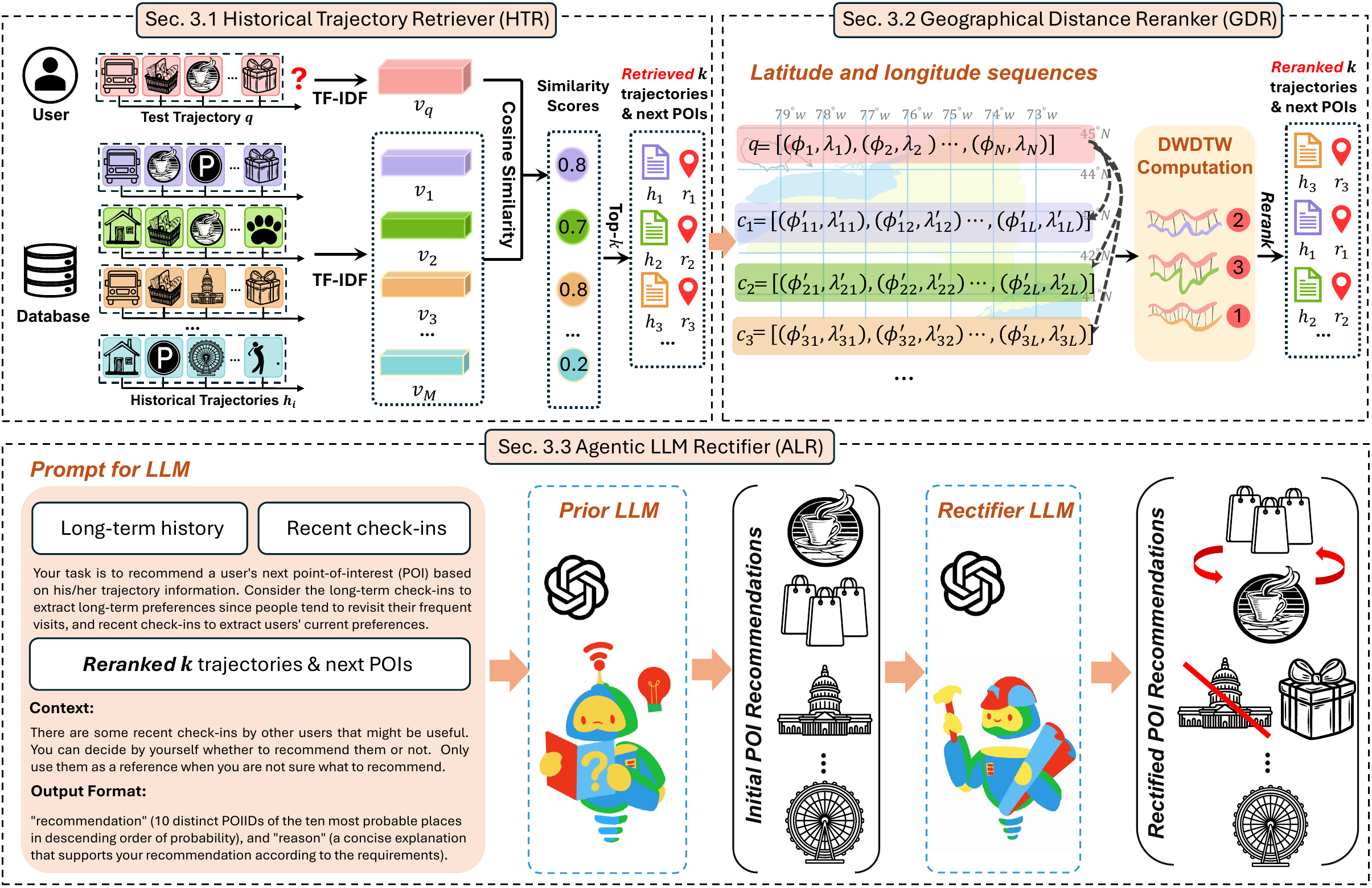}
  \caption{The overall framework of RALLM-POI for the next POI recommendation task.}
  \label{fig:main-placeholder}
\end{figure}

In this section, we present RALLM-POI, a retrieval-augmented LLM pipeline for next POI recommendation, as illustrated in Fig.~\ref{fig:main-placeholder}. Given a test trajectory, the Historical Trajectory Retriever (HTR) retrieves semantically relevant past trajectories and their recommendations as in-context examples. A Geographical Distance Reranker (GDR) then prioritizes those most aligned with the user’s recent spatial patterns, enhancing both behavioral and geographical relevance. Finally, the Agentic LLM Rectifier (ALR) acts as a quality assurance step, prompting another LLM to assess and refine the output for compliance and completeness. Together, these components enable RALLM-POI to deliver zero-shot recommendations that are both contextually informed and robustly validated.

\subsection{Historical Trajectory Retriever (HTR)}
To enhance the personalization and contextual relevance of recommendations, we propose Historical Trajectory Retriever (HTR), which augments each test trajectory with informative examples retrieved from a large database of historical user trajectories and their corresponding recommendations. The motivation behind this design is that similar behavioral histories often indicate similar user preferences, and grounding the prompt in authentic historical cases can lead to better guidance for LLMs during generation.

Formally, let $\mathcal{D} = \{(\mathbf{h}_i, \mathbf{r}_i)\}_{i=1}^M$ be the training database containing $M$ \textbf{historical trajectories} $\mathbf{h}_i$ of all users and their corresponding recommendations $\mathbf{r}_i$. Given a \textbf{test trajectory} $\mathbf{q}$, we first convert both $\mathbf{q}$ and each $\mathbf{h}_i$ into textual strings of location IDs. These strings are then projected into a vector space using an embedding function $\mathrm{Emb}(\cdot)$, which is instantiated as a TF-IDF encoder.
To identify the most relevant historical behaviors, we compute the cosine similarity between the embedding of the test user trajectory, $\mathbf{v}_q = \mathrm{Emb}(\mathbf{q})$, and each database embedding, $\mathbf{v}_i = \mathrm{Emb}(\mathbf{h}_i)$. The similarity score between $\mathbf{q}$ and the $i$-th historical trajectory is defined as $
\mathrm{sim}(\mathbf{v}_q, \mathbf{v}_i) = \frac{\mathbf{v}_q \cdot \mathbf{v}_i}{|\mathbf{v}_q| , |\mathbf{v}_i|}.$
The $k$ trajectories with the highest similarities are selected to form a supporting context. Formally, let $\mathcal{I}^*$ be the indices of the top-$k$ most similar trajectories:
\begin{equation}
\mathcal{I}^{*} = \underset{\mathcal{I}: |\mathcal{I}|=k}{\arg\max} \sum_{i \in \mathcal{I}} \mathrm{sim}(\mathbf{v}_q, \mathbf{v}_i)
\end{equation}
For each selected index $i \in \mathcal{I}^{*}$, the associated historical trajectories $\mathbf{h}_i$ are retrieved and combined with their corresponding POI recommendation $\mathbf{r}_i$ is denoted $c_i$, i.e., $c_i=\{\mathbf{h}_i,\mathbf{r}_i\}, i\in[1,k]$. 
The proposed HTR retrieves highly similar historical cases, providing the LLM with clear, contextually relevant guidance while avoiding distractors. Exposing these histories and their recommendations in the prompt helps align outputs with successful reference trajectories. To further improve quality, the retrieved trajectories are reranked before being used in the LLM prompt (see Subsection \ref{sec3.2}).

\subsection{Geographical Distance Reranker (GDR)}
\label{sec3.2}
After retrieving trajectories with HTR, we rerank them for geographical plausibility. While HTR identifies textually similar cases, they may not align spatially with the test trajectory. Since LLMs are sensitive to the quality and order of support examples \cite{jayaram2024mitigatingsensitivity2}, the Geographical Distance Reranker (GDR) prioritizes trajectories that best match the user’s recent spatial behavior.

Concretely, let the user’s recent trajectory of length $N$ be $\mathbf{q} = \{q_n\}_{n=1}^N$, where each $q_n$ is a visited POI mapped to a latitude and longitude pair $x_n = (\phi_n, \lambda_n)$. For the $i$-th candidate support trajectory $c_i = \{c_{il}\}_{l=1}^{L_{i}}$ retrieved by HTR, we similarly obtain the latitude and longitude pair sequence $y_{il} = (\phi_{il}', \lambda_{il}'),i\in[1,k],l\in[1,L_{i}]$. To measure spatial alignment between the user trajectory and each candidate, we propose a Decaying Weighted Dynamic Time Warping (DWDTW) distance based on real-world geospatial separation. Since trajectories vary in length and are often temporally misaligned, traditional point-to-point metrics like Euclidean distance could not appropriately handle these shifts. DTW \cite{jeong2011weighted} instead finds an optimal alignment between sequences by minimizing cumulative spatial cost on matched points.

Specially, to emphasize recent visited locations as they are more predictive of immediate intent, we implement a recency bias with exponentially decaying weights, $\omega_n = \rho^{N-n}$ for $\rho \in (0,1)$ and user POI length $N$.
The DWDTW alignment cost between $\mathbf{q}$ and a candidate $c$ is thus computed as:\begin{equation}
\text{DWDTW}(\mathbf{q}, c_i) = \min \sum_{n\in[1,N],l\in[1,L_{i}]} \omega_n \cdot d(x_n, y_{il}),
\end{equation}
where $d$ is the Haversine distance, $x_n$ is the user's recent trajectory, $y_{il}$ is the retrieved $i$-th trajectory. 
Lower WDTW values indicate stronger spatial consistency, especially at recent steps. DWDTW allows us to identify supportive trajectories even when users take slightly different paths. We compute DWDTW for all retrieved candidates $c_i$, rerank them in ascending order, and include them as contextual support in the LLM prompt. The LLM is also instructed to use its judgment in leveraging this context, avoiding over-reliance on retrieved information. Our proposed GDR injects spatial awareness into prompt construction, ensuring the most geographically plausible examples are prioritized and guiding the LLM to generate POI recommendations that reflect realistic movements.

\subsection{Agentic LLM Rectifier (ALR)}

To enhance reliability and compliance, we introduce an Agentic LLM Rectifier (ALR) inspired by recent agentic RAG work~\cite{singh2025agenticrag}. While earlier components retrieve and contextualize trajectories, structured recommendation tasks often require an extra verification to enforce constraints such as formatting, uniqueness, completeness, and intent alignment. Since LLMs in a single pass may overlook details, ALR adopts an agentic approach where the model self-assesses and revises its outputs, leveraging iterative reasoning to ensure more robust and reliable recommendations. Specifically, the agentic rectification prompt is structured as:

\begin{tcolorbox}[colback=gray!10!white, colframe=gray!80!black, title=Agentic LLM Rectifier Prompt]
A user has completed the following task:\\
<Prompt of prior LLM>\\
<Response of prior LLM>\\
\textbf{Your task:}\\
Carefully review the answer above.\
Evaluate whether it satisfies all requirements.\
If it is correct, concise, and well-formatted, reproduce the answer.\
If you find any issues (formatting, duplicates, insufficient recommendations, irrelevant POI IDs, unsatisfactory reasoning, etc), revise the answer to fully meet the requirements.\
\end{tcolorbox}

Therefore, the agentic rectifier LLM strategy serves as a quality assurance layer that systematically detects and amends errors that may otherwise propagate to the final output, and the final response strictly adheres to all constraints.

\section{Experiments}

\subsection{Overall Performance}
We conduct experiments on three Foursquare datasets \cite{foursquare} representing Singapore (SIN), New York City (NYC), and Phoenix (PHO). For preprocessing and ensuring data quality, POIs with fewer than 10 interactions are filtered out. We also remove users that have less than 5 trajectories and discard trajectories with no more than 3 POIs. The interaction records are chronologically sorted for each user, and 80\% of the data is used for database construction, while the remaining is for testing. Since our method is zero-shot, no validation set is required. RALLM-POI is compared with several representative training-based approaches and zero-shot LLM methods: ALSTM \cite{huang2019attention}, MCRNN \cite{liao2018mrnn}, CTLE \cite{lin2021clte}, CFPRec \cite{zhang2022cfprec}, LLMMob \cite{llmmob}, LLMMove \cite{feng2024llmmove}, and LLMZS \cite{llmzs}. For evaluation, two widely used metrics are employed: Hit Ratio (HR) and Normalized Discounted Cumulative Gain (NDCG). GPT-4o-mini is applied as the baseline LLM model for all zero-shot methods due to its capability and efficiency. The $\rho$ is set to 0.8 and we retrieve $k$=10 historical trajectories.

Table~\ref{tab:performance} summarizes the performance comparison across three datasets. 
Our method demonstrates strong and consistent performance across almost all datasets and metrics, which achieves especially high scores on the PHO dataset. Compared to traditional sequence-based models (ALSTM, MCRNN), which struggle to capture diverse and sparse locations, our approach integrates both contextual and geographical cues, yielding significantly higher hit rates and NDCG scores. Transformer-based models (CTLE, CFPRec) improve over sequential methods via attention, yet our framework consistently outperforms them on PHO and SIN datasets, demonstrating the superior generalization of LLMs enhanced by retrieval and reranking. Unlike zero-shot LLM-based methods (LLMMob, LLMMove, LLMZS) that rely on potentially irrelevant prompt contexts, our RAG based framework ensures high-quality, contextually relevant trajectory references, while the GDR further sharpens spatial relevance. Together, RALLM-POI can fully exploit LLM reasoning and generalization capabilities.

\begin{table*}[t]
    \caption{Performance comparison in HR@K and NDCG@K on three datasets, ”H” stands for HR, and ”N” stands for NDCG. The methods are grouped into training-based and zero-shot methods. Best results in bold and second-best in italics.}
\centering
\renewcommand{\arraystretch}{1.1}
\begin{tabular}{l|l|cccc|cccc}
\hline
\multirow{2}{*}{Data} & \multirow{2}{*}{Metric} & \multicolumn{4}{c|}{\textit{Training-based Methods}} & \multicolumn{4}{c}{\textit{Zero-shot Methods}} \\
 &    & ALSTM & MCRNN & CTLE & CFPRec & LLMMob & LLMMove & LLMZS & \textbf{Ours} \\
\hline

\multirow{4}{*}{PHO}
    & H@5  &           0.1579 & 0.1905 & 0.2632 & \textit{0.3421} &0.3158&0.3157 & \textit{0.3421}& \textbf{0.5263}\\
    & H@10 &           0.2377 & 0.2726 & 0.3605 & 0.4253 &0.5789&\textit{0.6051} & 0.5526 & \textbf{0.6842}\\
    & N@5  &           0.1033 & 0.1264 & 0.1995 & 0.2432 &0.2355& 0.2479& \textit{0.2520}& \textbf{0.3505}\\
    & N@10 &           0.1385 & 0.1617 & 0.2068 & 0.2730 &0.3193&\textit{0.3434} & 0.3191 & \textbf{0.3989}\\
\hline
\multirow{4}{*}{NYC}
    & H@5  &   0.1667 & 0.1835 & 0.2421 & \textbf{0.2734} & 0.2465&0.2470& 0.2020& \textit{0.2554}\\
    & H@10 & 0.2031 & 0.2397 & 0.3205 & 0.3306 &\textit{0.4071} &0.4023 &0.3723& \textbf{0.4190}\\
    & N@5  &  0.0912 & 0.1036 & 0.1513 & \textit{0.1588} & 0.1571&0.1565& 0.1285& \textbf{0.1599}\\
    & N@10 &  0.1638 & 0.1870 & 0.1841 & 0.1834 & \textit{0.2107}&\textbf{0.2132}& 0.1838& \textbf{0.2132}\\
\hline
\multirow{4}{*}{SIN}    
    & H@5  & 0.1296 & 0.1608 & 0.2041 & 0.2650 & \textit{0.2832}&0.2508 & 0.2785&\textbf{0.3148}\\
    & H@10 & 0.1933 & 0.1862 & 0.2784 & 0.3085 & 0.4240&0.2863& \textit{0.4335}& \textbf{0.4399}\\
    & N@5  & 0.1027 & 0.1169 & 0.1315 & 0.1588 & \textit{0.1957}&0.1455& 0.1924& \textbf{0.2231}\\
    & N@10 & 0.1476 & 0.1591 & 0.1556 & 0.1825 & 0.2414&0.1713& \textit{0.2422}&\textbf{0.2639}\\
\hline
\end{tabular}
\label{tab:performance}
\end{table*}

\begin{figure*}[t]
    \centering
    \includegraphics[width=0.99\linewidth]{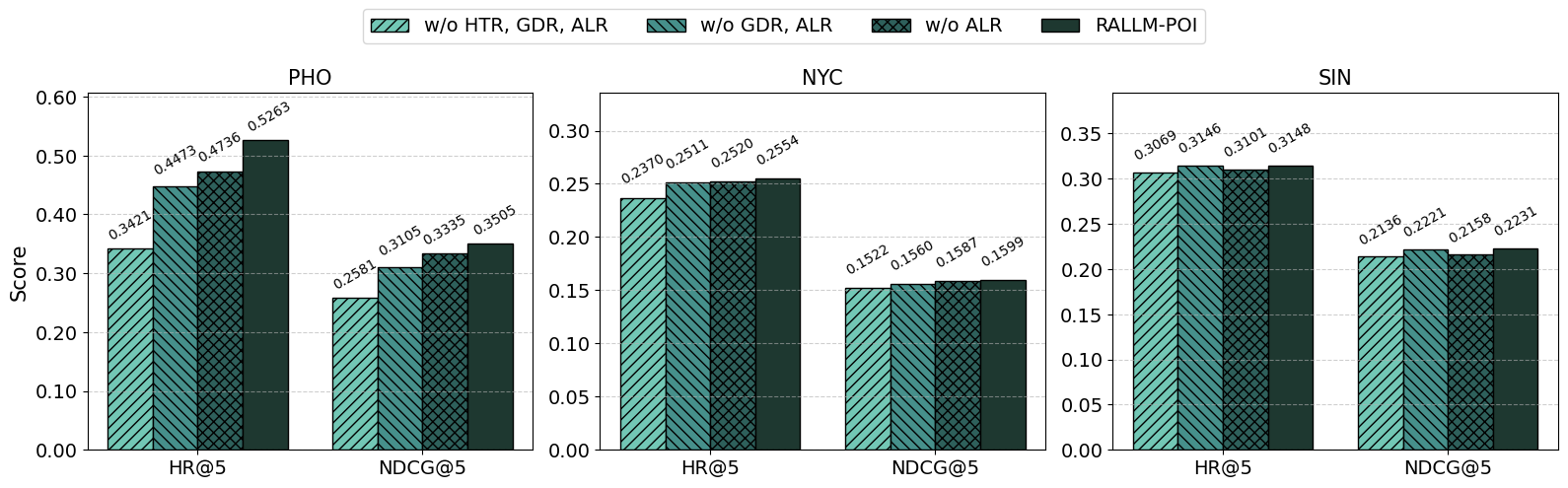}
    \caption{Ablation study of the proposed components HTR, GDR, and ALR.}
    \label{fig:ablation}
    \vspace{-12pt}
\end{figure*}

\vspace{-8pt}
\subsection{Ablation Study}

\textbf{Analysis on the proposed components.} We conduct an ablation study to evaluate each component of RALLM-POI (Fig.~\ref{fig:ablation}). The full model achieves the best HR@5 and NDCG@5 across all datasets, demonstrating the effectiveness of our design. Removing ALR (‘w/o ALR’) consistently lowers performance, highlighting its role in rectifying LLM outputs. Further removing GDR (‘w/o GDR, ALR’) reduces PHO HR@5 from 0.4736 to 0.4473 and NDCG@5 from 0.3335 to 0.3105, showing its importance in leveraging historical geographical context. When all three components are removed (‘w/o HTR, GDR, ALR’), performance drops most significantly (e.g., NYC HR@5: 0.2370, NDCG@5: 0.1522). These results indicate that each component contributes both independently and cumulatively to RALLM-POI’s effectiveness.

\begin{figure*}[t]
    \centering
    \includegraphics[width=0.99\linewidth]{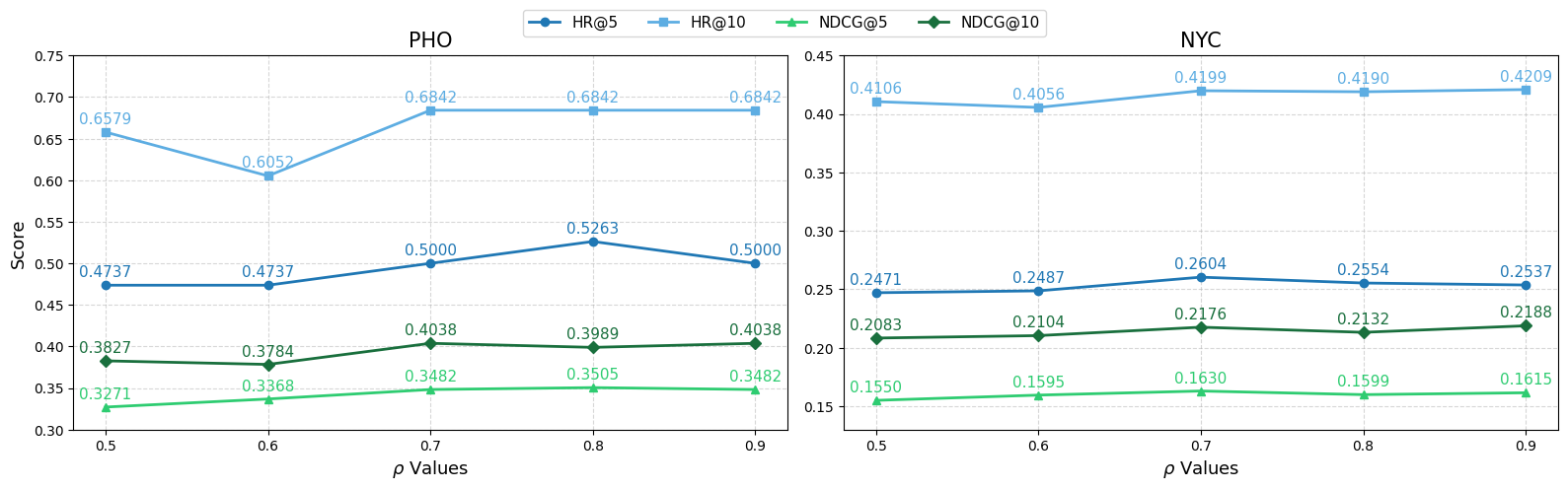}
\vspace{-10pt}
    \caption{Performance comparison with different decaying weight $\rho$ in DWDTW.}
    \label{fig:lambda}
\vspace{-4pt}
\end{figure*}

\noindent\textbf{Analysis on the decaying weight $\rho$ of DTW.}
In GDR, $\rho$ controls the spatial alignment cost so that recent locations in a trajectory receive greater weight in DWDTW. As shown in Fig.~\ref{fig:lambda}, increasing $\rho$ from 0.5 to 0.8 improves PHO performance (HR@5: 0.4737 to 0.5263, NDCG@5: 0.3271 to 0.3505), highlighting the benefit of prioritizing recent points. However, setting $\rho$ too high (e.g., 0.9) reduces accuracy, likely due to overemphasis on recency at the expense of contextual diversity. A similar trend is observed on NYC, where performance peaks at $\rho=0.7$–0.8. We therefore adopt $\rho=0.8$ as a balanced choice, yielding robust gains while preserving spatial breadth.

\begin{table*}[t]
\centering
\renewcommand{\arraystretch}{1.0} 
\setlength{\tabcolsep}{6.5pt} 
\caption{The performance of different user groups for user cold-start analysis.}
\begin{tabular}{lcccccc}
\hline
{User} & \multicolumn{2}{c}{\textbf{PHO}} & \multicolumn{2}{c}{\textbf{NYC}} & \multicolumn{2}{c}{\textbf{SIN}} \\
\cline{2-3} \cline{4-5} \cline{6-7}
{Groups} & {HR@5} & {NDCG@5} & {HR@5} & {NDCG@5} & {HR@5} & {NDCG@5} \\
\hline
Inactive       & 0.5000 & 0.3874 & 0.2928 & 0.1788 & 0.2990 & 0.2026 \\
Normal         & 0.5333 & 0.3644 & 0.2863 & 0.1903 & 0.3083 & 0.2230 \\
Very Active    & 0.4745 & 0.2680 & 0.1807 & 0.1058 & 0.3249 & 0.2136 \\
\hline
\end{tabular}
\vspace{-15pt}
\label{tab:user_group_performance}
\end{table*}

\noindent\textbf{Analysis on user cold-start performance.}
The cold-start problem arises when inactive users provide limited trajectories, making their patterns harder to capture \cite{Li_2024}. To evaluate adaptability, we divide users into inactive (bottom 30\%), normal, and very active (top 30\%) groups based on long-term POIs. Results on PHO, NYC, and SIN (Table~\ref{tab:user_group_performance}) show that inactive users often achieve the highest performance, e.g., PHO with NDCG@5 of 0.3874 and NYC with HR@5 of 0.2928, outperforming very active users. This trend reflects the strength of our HTR module, which retrieves trajectories from similar users to enrich sparse histories, effectively mitigating cold-start challenges.

\vspace{-8pt}
\section{Conclusion}
\vspace{-8pt}
This paper presents RALLM-POI, a zero-shot next POI recommendation framework that combines LLM generalization with retrieval-augmented generation and self-rectification. Unlike traditional training-based methods, RALLM-POI leverages LLM knowledge, enriched with contextually and geographically relevant trajectories via HTR and refined by GDR, while ALR provides self-evaluation and correction for quality assurance. Evaluations on three real-world datasets show substantial gains over both training-based and state-of-the-art zero-shot LLM methods. Ablation studies highlight how each component synergistically enhances accuracy and robustness, particularly addressing cold-start scenarios while maintaining adaptability across diverse user trajectories.

{
\vspace{2mm}
\footnotesize
\noindent{\textbf{Acknowledgements}}.
This research is supported by the Ministry of Education, Singapore (MOE), under its Academic Research Fund Tier 2 (Award No. MOE-T2EP20123-0015). Any opinions, findings and conclusions, or recommendations expressed in this material are those of the authors and do not reflect the views of MOE.
}

%
%
%
\bibliographystyle{splncs04}
\bibliography{mybib}
\end{document}